\documentclass[sigconf]{acmart}

\AtBeginDocument{%
  \providecommand\BibTeX{{%
    \normalfont B\kern-0.5em{\scshape i\kern-0.25em b}\kern-0.8em\TeX}}}

\setcopyright{acmcopyright}
\copyrightyear{2022}
\acmYear{2022}

\acmConference[CIKM '22]{Human-in-the-Loop Data Curation Workshop at the 31st ACM International Conference on Information and Knowledge Management}{October 17--22, 2022}{Atlanta, GA}

\begin{document}

\title{A Human-ML Collaboration Framework for Improving Video Content Reviews}

\author{Meghana Deodhar, Xiao Ma, Yixin Cai, Alex Koes, Alex Beutel, Jilin Chen}
\affiliation {
    \institution{Google}
    \country{USA}
}
\email{{mdeodhar, xmaa, yixincai, koes, alexbeutel, jilinc}@google.com}

\begin{abstract}

We deal with the problem of localized in-video taxonomic human annotation in the video content moderation domain, where the goal is to identify video segments that violate granular policies, e.g., community guidelines on an online video platform. High quality human labeling is critical for enforcement in content moderation. This is challenging due to the problem of information overload - raters need to apply a large taxonomy of granular policy violations with ambiguous definitions, within a limited review duration to relatively long videos. Our key contribution is a novel human-machine learning (ML) collaboration framework aimed at maximizing the quality and efficiency of human decisions in this setting - human labels are used to train segment-level models, the predictions of which are displayed as “hints” to human raters, indicating probable regions of the video with specific policy violations. The human verified/corrected segment labels can help refine the model further, hence creating a human-ML positive feedback loop. Experiments show improved human video moderation decision quality, and efficiency through more granular annotations submitted within a similar review duration, which enable a 5-8\% AUC improvement in the hint generation models.

\end{abstract}


\maketitle

\section{Introduction}
\label{content-moderation}

The importance of content moderation on online video platforms such as TikTok, YouTube or Instagram is growing %
\cite{meta-hs-misinfo,tiktok-misinfo}. These platforms strive to accurately detect the presence of policy violations within the video, which drive enforcement actions, e.g., the video can be taken down.
Given the complexity of this problem, content moderation relies heavily on human judgement and employs large teams of content moderators to perform reviews. Since human annotations directly lead to high stakes decisions, such as content take downs,
the quality of the annotations is critical.

For content moderation decisions there is a growing need for transparency in detected policy violations to provide feedback to content creators \cite{jhaver2019}.
This motivates a in-video taxonomic annotation task, where the goal is to provide \emph{localized} and \emph{fine-grained policy-specific} annotations, i.e., both the time regions (video segments) and the exact policies violated, which inform downstream video content moderation decisions. To cover the spectrum of potential violations, policy playbooks typically contain hundreds of fine-grained policies. This large space of policy violations can be organized as a taxonomy of broad categories such as Profanity, Violence, Nudity, etc., each of which contains several granular violations. For instance, Violence could include a range of granular classes such as animal abuse or graphic violence in video games.
The class definitions are complex, ambiguous and often require nuanced judgment to apply, e.g., graphic violence. New policy classes may be added over time as well, e.g., Covid anti-vaccination.
Moreover, there is a class imbalance issue - some egregious violations may be very rare.

Our goal is to maximize the quality and efficiency of the complex, granular, localized policy annotations task, hence leading to the correct video level enforcement decision. We achieve this by tackling the key issue of "information overload" faced by raters in providing high quality annotations, where 1) the sheer volume of videos on large online video platforms means raters only have limited review time per video; and 2) the large taxonomy of policies makes it hard for raters
to recall the complete set of all granular violations for every video region they watch in the limited review duration.

We propose a human-ML collaboration framework to maximize human ratings quality and efficiency by addressing "information overload". We train models on granular rater annotations to predict policy violations, which are then combined with innovative front-end elements in the rating tool to provide "hints" to assist raters. 
We borrow from information retrieval literature and use ranking mechanisms for identifying the most useful and succinct set of hints. In experiments, we show that this enables raters to efficiently label policy violations more correctly and comprehensively. The human interactions with model hints pave the way for leveraging human feedback to improve the underlying ML models.

\section{Related Work}

The crowd sourcing literature is very rich in the application of human annotations to perform a variety of tasks such as text processing \cite{kittur2008,Bernstein2010,Hu2011}, audio transcription \cite{Lasecki2012}, taxonomy creation \cite{chilton2013cascade}, and social media analysis~\cite{zubiaga2015crowdsourcing,founta2018large}. Although there is existing literature on video annotation, it is primarily focused on identifying actions or labeling entities easily distinguishable by humans using visual information only, e.g., high jump, thunderstorms. The primary goal of these tasks is to create large datasets for facilitating Machine Learning/Perception applications \cite{Zhao2017, trujillo2022mela}, e.g., the YouTube-8m dataset \cite{Haija2016}. This is very different from our set up, where raters annotate granular, ambiguously defined policies using multi-modal signals - video, audio and text, from the transcript, and the video title and description. The recent emergence of crowd sourcing literature on content moderation primarily covers textual content such as user comments \cite{Lai2022,Chandrasekharan2019}, however there has been little focus on video moderation tasks.

Human-ML collaboration is an emerging area of research with two main categories of work:

\textbf {ML-Assisted Human Labeling}: ML-assistance through predictions and explanations has been used to improve the quality of human decisions in several domains \cite{Beede2020,Lai2019,Park2019}, including content moderation \cite{Chandrasekharan2019,Lai2022}. Crossmod \cite{Chandrasekharan2019}, for instance, uses a model trained on historic cross-community moderation decisions to enable Reddit human moderators to find more violations. Interactive ML-assistance, which we leverage in Section \ref{pre-populated_segments}, is used by Bartolo et al. \cite{Bartolo2022} to assist human annotators to develop adversarial examples for improving a natural language question answering model. ML-assistance has been shown to also improve the efficiency of the human labeling task \cite{Ashktorab2021, Chandrasekharan2019,Anjun2021}. Our work aims to exploit both the human labeling quality and efficiency benefits.

\textbf {Improving ML-models Through Human Annotations}: Human annotations are useful in constructing hybrid human-ML systems that leverage the complementary strengths of both to improve the performance of ML models \cite{Arous2020,Vaughan2018}. Existing work on active learning~\cite{settles2009active, yang2015multi} shows that strategically sampling data points can reduce human workload, but the purpose is to improve machine learning models instead of assisting raters.
Recent work on explainable active learning (XAL)~\cite{Ghai2020} has called for better designing for the human experience in the human-AI interface.

Our work shows that it is possible to achieve model improvements and assist human raters, bridging the gap between ML-assisted human labeling and active learning.
The novelty of our approach is that the models are re-trained continuously on the output of the human annotation task, which they provide assistance for, constructing a positive feedback loop between humans and models. In this collaborative framework, we have the opportunity to improve both modeling and human rater performance.

Information overload, which we encounter in our content moderation setting, is a well studied problem that reduces the effectiveness of a human's decision making ability~\cite{eppler2008concept,chen2009effects}.
To address this, we build on the intuition that humans find it easier to verify or correct suggestions rather than produce new annotations from scratch \cite{Hu2020,Bernstein2010}.
Our ML-assistance proposal strives to select the most informative but succinct ML-based "hints" to surface to raters by drawing on the information retrieval and ranking literature. ML-based ranking has been shown to reduce information overload effectively in electronic messaging \cite{losee1989minimizing}  and social media~\cite{koroleva2012reducing,chen2010short}.
We draw inspiration from the learning to rank idea~\cite{karatzoglou2013learning} to reduce information overload for raters.

\section{Proposed Human-ML Collaboration Framework}

\begin{figure}[h]
  \centering
  \includegraphics[width=\linewidth]{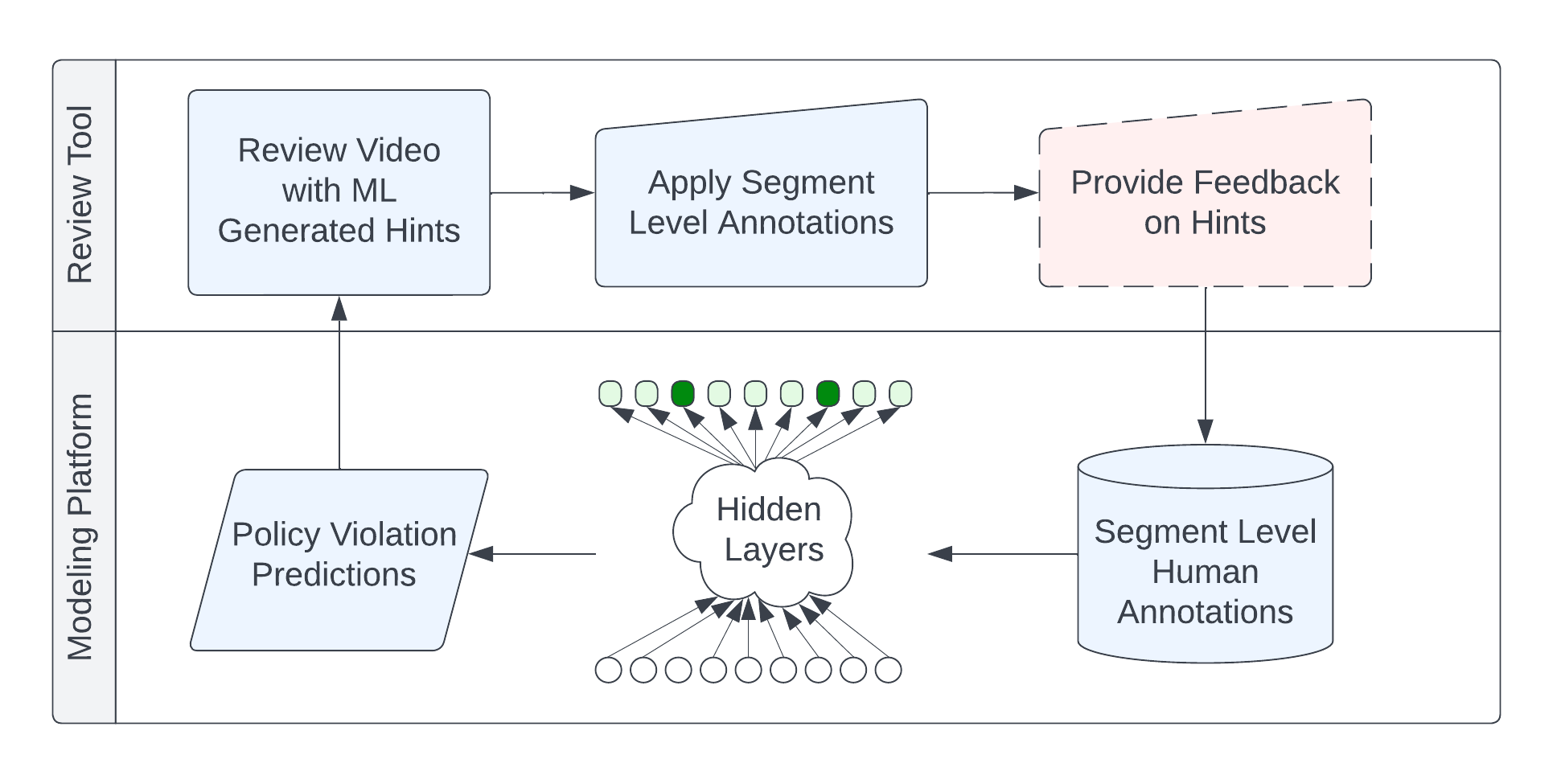}
  \setlength{\abovecaptionskip}{-9pt}
  \setlength{\belowcaptionskip}{-9pt} 
  \caption{Human-ML collaboration set up.}
  \Description{Human-ML collaboration set up.}
  \label{fig:human-ml}
\end{figure}

The main contribution of this paper is the human-ML collaboration framework visualized in Figure \ref{fig:human-ml}. We use the predictions of ML models to provide assistance to human raters and evaluate the effectiveness of different user interfaces for the ML-assistance.
Since the policy violation prediction task is hard for ML models, the feedback from human raters is useful to improve the models. ML-assistance enables raters to provide segment level annotations more efficiently leading to more ground truth to train/update the ML models. Additionally, we can enable raters to interact with the ML hints (accept/reject), providing direct feedback to refine the model, establishing a positive human-ML feedback loop.

\subsection{ML-Assisted Human Reviews}

As discussed in the introduction, raters face an information overload problem due to a combination of granular, complex policy definitions and limited time per video.
Often, raters need to use their judgement to decide what parts of the video to watch to identify violating segments, resulting in fleeting violations being missed, or inconsistencies across raters watching different sections of the same video.

It is intuitive that raters would benefit from pointers to likely unsafe regions within a video, labeled with the exact policies being violated. Even if not completely precise, this will enable them to optimize their review bandwidth by focusing on potentially more relevant regions, making them less likely to miss violations. We achieve this by training per policy ML models and transforming their predictions into "hints" provided to raters, described in more detail in Section \ref{pre-populated_segments}.
We tune the models to be high recall to minimize uncaught violations, while relying on human judgement to improve the precision of the labeled violations.

\subsubsection{Segment Level Model Training}

To train segment level policy violation models we frame the following modeling problem - given multi-modal features for a fixed length video segment, predict whether the segment contains specific policy violations. We generate training datasets by extracting per-frame visual and audio features from the human labeled violation region. The visual and audio features are the dense embedding layers of standard convolutional network image semantic similarity \cite{Szegedy2015} and audio classification models \cite{Hershey2016} respectively. Based on empirical evidence, we select flat concatenation to aggregate the frame features over a segment, versus average/max pooling. The final model we train is a multi-label DNN model with the aggregated frame-level visual and audio features as input, where each label corresponds to a fine grained policy violation.  We use MultiModal Versatile Networks \cite{alayrac2020self} during model training to learn a better representation for audio and visual features for our classification task, which further improves model performance. We use a sliding window approach to utilize the trained model to generate prediction scores per policy violation for a fixed length segment starting at each frame of the video.  Using a window of frame size n and stride of 1 frame, we produce model scores for segments with start and end frames [0 to n-1], [1 to n], and so on until the end of the video, padding with empty features to fit the segment length for the last n frames.

\subsubsection{Techniques to Provide ML-Assistance}
\label{pre-populated_segments}

We proceed to develop ways to use model predictions to most effectively assist human reviews, and provide details on two different designs (V1 and V2).

\begin{figure}[h]
  \centering
  \includegraphics[width=\linewidth]{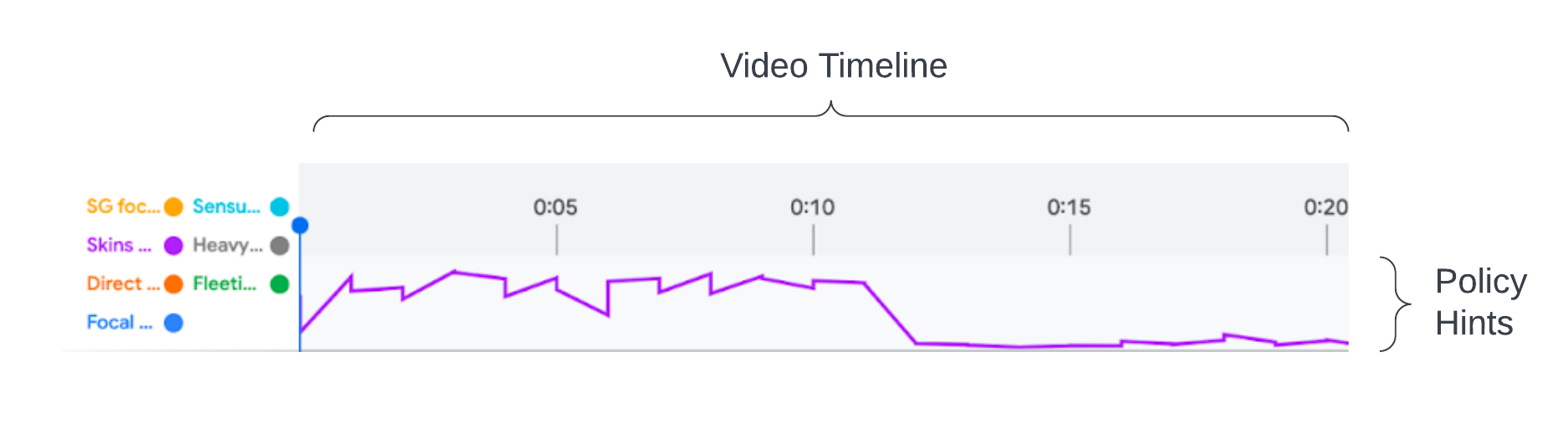}
  \setlength{\abovecaptionskip}{-9pt}
  \setlength{\belowcaptionskip}{-9pt} 
  \caption{V1 Continuous Line Graph for Specific Policy Hints}
  \Description{Line Graph Hints}
  \label{fig:v1_UI}
\end{figure}

\textbf{V1 Hints: Continuous Line Graphs.}
For video annotations, it is standard to display the video itself with playback controls and additional information in the form of a timeline~\cite{vondrick2010efficiently}.
For V1, we display the ML predictions as a line graph across the entire timeline of the video. The user interface is demonstrated in Figure \ref{fig:v1_UI}.
Raters can examine the line graphs and jump to the point in the video where a peak (policy violation) occurs.

While we have model predictions for hundreds of granular policies, due to visual clutter, we only display plots for a small subset of the most frequent policies.
Raters also don't have the ability to provide feedback to improve model predictions.

\begin{figure}[h]
  \centering
  \includegraphics[width=\linewidth]{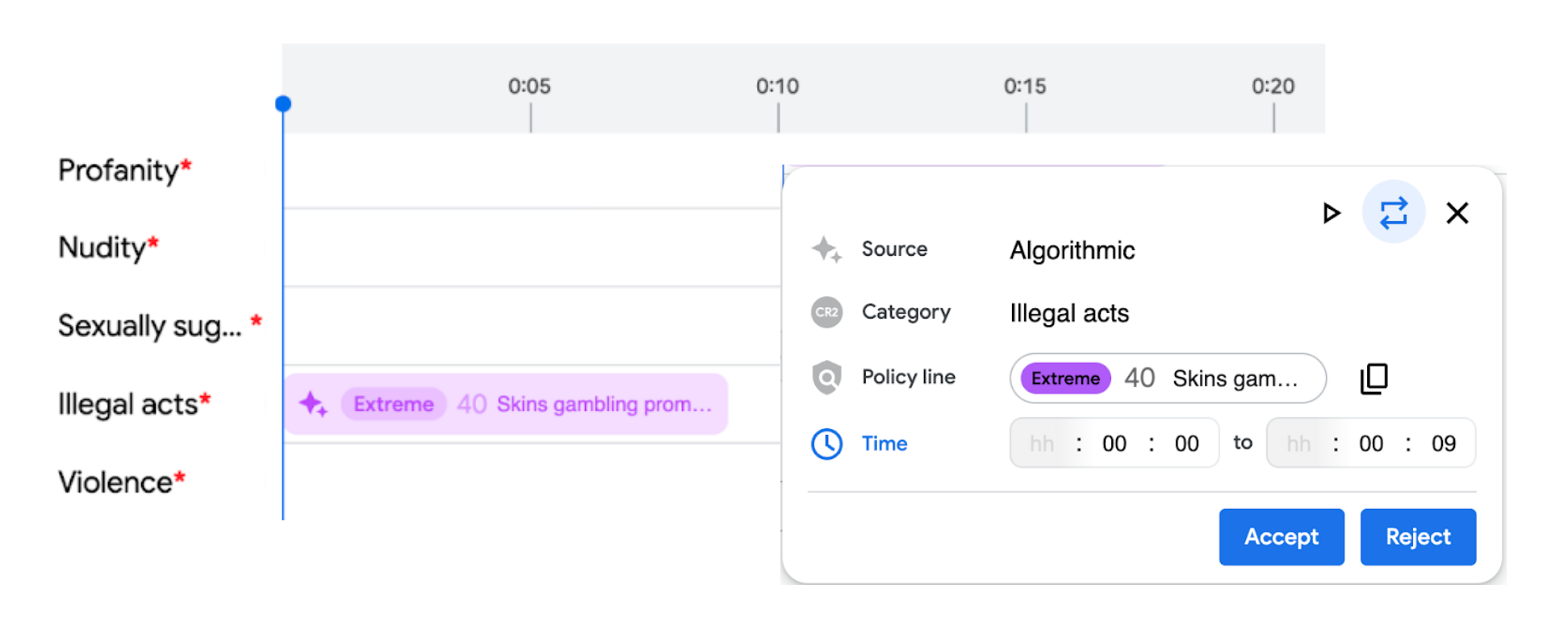}
  \setlength{\abovecaptionskip}{-9pt}
  \setlength{\belowcaptionskip}{-9pt} 
  \caption{V2 Pre-Populated Segments from ML Models}
  \Description{Pre-Populated Segments}
  \label{fig:v2_UI}
\end{figure}

\textbf{V2 Hints: Towards a Scalable and Interactive-ML Assistance UI.}
In V2, we borrow elements from recommender systems~\cite{portugal2018use,swearingen2002interaction} to develop a more \emph{scalable} and \emph{interactive} interface (see Figure \ref{fig:v2_UI}), where we pre-populate video segments that may contain policy violations detected by machine learning models.

To generate video segments, as shown in Figure \ref{fig:v2_UI}, we introduce an algorithm to binarize continuous model scores per policy into discrete segments and use a ranking algorithm to recommend the most useful segments to raters.
For simplicity, we used a threshold-based algorithm.
The threshold selection constitutes a tradeoff between the precision and recall, where precision captures the utility of the predicted segments to raters and recall captures the comprehensive coverage of all video violations. The algorithm chooses a threshold maximizing recall, while maintaining a minimum precision (40\% based on user studies). The regions of the video where the ML scores are above this chosen threshold are displayed as predicted policy violating segments to raters. Several heuristics are then applied to maximize segment quality, e.g., we merge segments that are close to avoid visual clutter (<3\% of the whole video length apart).

Finally, to reduce information overload, we borrow from the learning to rank concept \cite{karatzoglou2013learning} to rank candidate segments and limit the number of displayed segments.
The ranking algorithm prioritizes segments based on the max score across segment frames, and egregiousness of the predicted policies.
We then select the top $N$ segments to display to raters, with $N$ selected through user studies.
We pre-populate each ML suggested segment in the video timeline UI as seen in Figure \ref{fig:v2_UI}.
Raters can choose to accept or reject the suggested segments.
These logged interactions can be used to provide feedback to improve the ML models.

\subsection{Human Feedback to ML-Models}

The ambiguous and fluid policy definitions along with a changing distribution of videos on online platforms poses a challenge for building robust models to accurately predict policy violations for providing ML-assistance. We hence continuously need more data and human feedback to improve the models.
We show that ML-assistance increases the total number of segment labels submitted vs. no assistance. The labels in turn can serve as new ground truth for continuously re-training the models and improving performance.
Additionally, with the reject button shown in V2 we collect clean negative labels; earlier, we had only "weak" negatives from videos where no violations were annotated \footnote{Even if no violations were annotated, they could still be present in video segments the rater did not watch, hence the negative labels inferred are weak/noisy.}. Further exploiting the human feedback in combination with active learning strategies is an area of future work.

\section{Experiments}

Our proposed methodology is evaluated using raters from the author's organization that regularly perform video content moderation reviews on live queues. Raters are separated into two pools: experts and generalists, with 150 and 400 raters respectively. The expert pool is a group of more experienced quality assurance (QA) raters with demonstrably better decision quality over a long period of time. Since their focus is QA, they don't have fixed productivity targets as generalists do. They can hence spend more time reviewing each video comprehensively, leading to higher decision quality.
In our experiment setup, each video in an evaluation dataset is independently reviewed by 1 expert and 2 generalist raters. We use the labels from expert raters on the dataset as the ground truth to evaluate the 2 sets of generalist rater decisions.

\subsection{Experimental Setup}

\subsubsection{Datasets}
Our two evaluation datasets contain videos viewed on a large online video platform:
(1) "Live traffic" dataset: Sampled from live traffic, hence containing a very low proportion of policy violating videos;
(2) "Affected slice" dataset, sampled from live traffic and filtered to only videos with ML-hints present, containing 13-20\% policy violating videos.

\subsubsection{Metrics}
Our human ratings quality metric is calculated at the video-level and conveys the correctness of the final, binary content moderation decision, e.g., take down or not. We compute the precision (P), recall (R), and disagreement rate
for each of the 2 sets of generalist's video-level decisions. We consider the expert decision as ground truth and report the averaged values across the 2 sets.
On live traffic datasets, we use P/R over standard inter-rater disagreement metrics due to the high class imbalance.

For rater efficiency, we measure:
(1) percentage of policy violating videos where raters provide segment annotations.
(2) number of segment annotations submitted by raters per video
(3) average review duration per video.

\subsection{Results}

\subsubsection{Rater Quality Improvements}
We conduct experiments on both "live traffic" and "affected slice" datasets, with the baseline as a review process without ML-hints.
Tables~\ref{table1} and ~\ref{table2} compare the ratings quality metrics of our proposed V1 (line plot of model scores) ML-assistance treatment relative to the baseline, and evaluate the incremental benefit of the V2 (pre-populated segments) treatment over V1, in the V1 + V2 vs. V1 row.

\vspace{-1em}
\begin{table}[h]
\centering
\begin{tabular}{ | p{0.11\textwidth} | p{0.06\textwidth} | p{0.06\textwidth} | p{0.06\textwidth} | }
  \hline
  Treatment & Precision & Recall & \# Videos \\ 
  \hline
  V1 vs. Baseline & +9.82\% & +1.37\% & 3456 \\
  \hline
  V1 + V2 vs. V1 & +9.97\% & +5.64\% & 2914 \\
  \hline
\end{tabular}
\caption[Relative impact on live traffic (Dataset 1)]{Relative impact on live traffic (Dataset 1)}
\label{table1}
\end{table}

\vspace{-3.5em}
\begin{table}[h]
\centering
\begin{tabular}{ | p{0.11\textwidth} | p{0.06\textwidth} | p{0.05\textwidth} | p{0.11\textwidth} | p{0.06\textwidth} | }
  \hline
  Treatment & Precision & Recall & Disagreement\% & \# Videos  \\ 
  \hline
  V1 vs. Baseline & +4.24\% & +3.64\% & -15.71\% & 3319 \\
  \hline
  V1 + V2 vs. V1 & +7.02\% & +14.30\% & -32.27\% & 682 \\
  \hline
\end{tabular}
\caption{Relative impact on affected slice (Dataset 2)}
\label{table2}
\end{table}
\vspace{-2.5em}

From the live traffic results in Table~\ref{table1}, we see that V1 shows an improvement in precision and recall over the baseline, driven by the improvement on the affected slice as seen in Table~\ref{table2}.

We also see large rater quality gains of V2 over V1 on both live traffic and affected slice datasets. The segmentation and ranking algorithms in V2 allows us to overcome the scalability limitation of V1 and expand the number of granular policies covered by model hints from 7 to 18. Specifically for violence related violations, we see a 35\% relative recall gain over V1 by expanding policies with ML-hints from 2 to 9. The V2 design can be scaled to cover hundreds of policies in future versions by dynamically surfacing the most relevant violating segments, further improving recall.

\subsubsection{Rater Efficiency Improvements}
We observe reduced review duration on policy violating videos with V1 hints vs. without, with more efficiency benefits on longer videos as expected; -14\%  on videos longer than 10 minutes and -20\% on videos longer than 30 minutes. With V2, relative to V1, we see a 3\% increase in review duration, but it is traded off by a 9\% increase in the percentage of policy violating videos with exact segments annotated, and a 24\% increase in the number of segment annotations submitted per video.

\subsubsection{Interactive ML-Assistance Metrics}

Isolating the precision of the ML-assisted segments, we see raters accepting 35\% of ML generated hints, which is in line with the 40\% precision constraint we chose when converting model scores into discrete segments.

\subsubsection{Model Quality Improvements}
\label{model-quality-results}
Since the introduction of V1 hints, we see significant model performance improvements with more human labels collected within a 3 month period on specific policy areas. 

\vspace{-0.5em}
\begin{table}[h]
\centering
\begin{tabular}{ | p{0.14\textwidth} | p{0.10\textwidth} |p{0.12\textwidth} | }
  \hline
  Policy Area & AUCPR & \# Positive Labels \\ 
  \hline
  Sexually Suggestive & +5.9\% & +12.7\% \\
  \hline
  Nudity & +4.9\% & +12.7\% \\
  \hline
  Illegal Acts & +8.6\% & +12.1\% \\
  \hline
\end{tabular}
\caption{Model Quality Improvements}
\label{table3}
\end{table}
\vspace{-3em}

\section{Discussion}

One of the potential risks of our proposed human-ML collaboration framework is automation bias~\cite{skitka1999does}, where a rater's over-reliance on ML-assistance can result in (i) blind-spots due to humans missing violations and (ii) raters accepting model hints without verification. Our video-level ratings quality evaluation metrics are robust to this since the ground truth comes from expert (QA) raters who review videos comprehensively, looking beyond ML-hints. In practice, we observe little evidence of both (i) and (ii). 56\% of the violation segments submitted by raters in the V2 setup are organically created, i.e., don’t overlap with pre-populated hint segments. The segment acceptance rate is 35\%, aligned with our segmentation model precision tuning point of 40\%, indicating that raters are verifying and rejecting false positive hints at the expected rate. We could mitigate the risk of (ii) further by enforcing that at least some percentage of hint segments/video is actually watched or by surfacing the model's confidence in the predicted hint to raters. To ensure robust evaluation of model quality, the AUC improvements in Section~\ref{model-quality-results} are evaluated on a set of labels collected without model generated segments.

\section{Future work}
For content moderation to scale to the size of online platforms, it is necessary to take model-based enforcement action.
We would like to explore the relation between improved ground truth and improvement of automated, model based enforcement. Leveraging active learning strategies in combination with utilizing rater feedback on model generated segments to show further quality improvements in the models is another open area of research.
Finally, we will explore multi-armed bandits to balance active learning based \emph{exploration} for model improvement with model \emph{exploitation} for providing high quality ML-assistance~\cite{mcinerney2018explore}.

This paper used content moderation as the test bed for our human-ML collaboration proposal. However, it is a more generalized framework that applies to the problem of granular, localized video annotation encountered in various other industry applications such as identifying products/brands in videos to inform the placement of relevant ads, which we would like to explore further.

\bibliographystyle{ACM-Reference-Format}
\bibliography{sample-base}

\end{document}